\documentclass{article}
\usepackage{spconf,amsmath,graphicx}
\usepackage{times}
\usepackage{epsfig}
\usepackage{graphicx}
\usepackage{amsmath}
\usepackage{amssymb}
\usepackage{times}
\usepackage{subfigure}
\usepackage{hyperref}
\usepackage{gensymb}
\usepackage{xcolor}
\usepackage{bm}
\def\g{{\bm{g}}}

\def\G{{\bm G}}
\def\h{{\bm h}}

\def\etal{{\em et al.\/}}

\def\ie{{\em i.e.\/}}

\newcommand{\RNum}[1]{\uppercase\expandafter{\romannumeral #1\relax}}

\title{Multi-Resolution
Overlapping Stripes Network for Person Re-Identification}
\name{Arda Efe Okay$^{\dagger}$ Manal AlGhamdi$^{\star}$ 
Robert Westendorp$^{\ddagger}$
Mohamed Abdel-Mottaleb$^{\dagger}$}

\address{$^{\dagger}$ Department of Electrical and Computer Engineering, University of Miami, USA\\ $^{\star}$Department of Computer Science, Umm Al-Qura University, Saudi Arabia\\
$^{\ddagger}$ Fortinet Technologies ULC. , Canada }
\begin{document}
%
\maketitle
\begin{abstract}
This paper addresses the person re-identification (PReID) problem by combining global and local information at multiple feature resolutions with different loss functions.
Many previous studies address this problem using either part-based features or global features. 
In case of part-based representation, the spatial correlation between these parts is not considered, while global-based representation are not sensitive to spatial variations.
This paper presents a part-based model with a  multi-resolution
network that uses different level of features.
The output of the last two conv blocks is then partitioned horizontally and processed in pairs with overlapping stripes to cover the important information that might lie between parts.
We use different loss functions to combine local and global information for classification.
Experimental results on a benchmark dataset demonstrate that the presented method outperforms the state-of-the-art methods.

\end{abstract}
\section{Introduction}
\label{sec:intro}

Person re-identification (PReID) is the task of identifying the presence of a person from multiple surveillance cameras.
Given a query image, the aim is to retrieve all images of the specified person in a gallery dataset.
This task has attracted the attention of many researchers in computer vision for its great importance in multiple applications such as video surveillance for public security.
With the recent success of deep convolution neural networks (CNNs), PReID performance has made significant progress.
Deep representations provide high discriminative ability, especially when aggregated from part-based deep local features. 

Current related studies in PReID can be categorized to global feature-based and local part-based models. 
The local part-based models perform better with certain variations such as partial occlusion. 
Sun \etal \cite{Sun2018}, for instance, presented the part-based convolutional baseline (PCB) that horizontally divided the last feature maps into multiple stripes where each one contains part of the person's body in the input image. 
After that, a refinement mechanism was applied to each piece to guarantee that the feature map of this part focuses on the correct body part. 
PCB is a simple and effective framework that outperforms the other part-based models. 
However, it does not consider global features which play an important role in recognition and identification tasks and are normally robust to multiple variations. On the other hand, since their stripes have no overlaps, it loses important information that might lie at the edges of the divided stripes.

Global feature-based models focus on contour, shape, and texture representations.
For example, Wang \etal \cite{Wang2018} built the DaReNet model based only on global information
using a multiple granularity network to extract global features at different resolutions.
Hermans \etal \cite{Hermans2017} presented a ResNet-50 based classifier which uses global information.
Shen \etal \cite{GSRW} combined global features with random walk algorithm. 
Li \etal \cite{Li2018} proposed an attention-based model.
Luo \etal \cite{Luo_2019} reported a strong CNN based model with bag of learning tricks including augmentation and regularization. 
However, those methods may fail in the presence of object occlusion, multiple poses and lighting variations and usually depend on pre- and post-processing steps to boost their performances.

To address the above problems, other groups combined both global and local features.
Li \etal \cite {li2019pedestrian} fused local and global features while using mutual learning but they did not train the model with multiple loss functions.
While He \etal \cite{he2019mfbn} used attention aware model that combines global and local features.
Quan \etal \cite{Quan2019} introduced neural architecture search to PReID by focusing on searching the best CNN structure and applied the part-aware module in PReID search space that employs both part and global information.

\begin{figure*}
\centering
\includegraphics[width=0.9\textwidth]{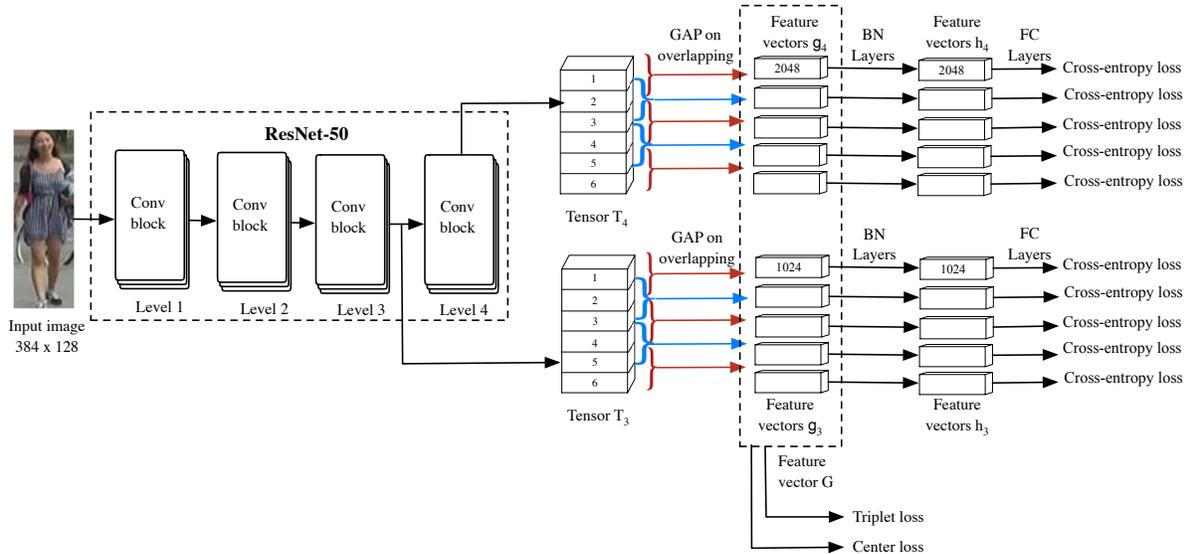}
\caption{Illustration of multi-resolution overlapping stripes (MROS) model for PReID. The backbone network is ResNet-50 \cite{Resnet} with four conv blocks. The input image goes through ResNet-50 to generate two feature tensors T$_j$ where $j= 3$ or $4$, which are used by the overlapping stripe approach to obtain $\g_j$ vectors. 
The global feature vector $\G$ is then obtained by as a concatenation of $\g_j$ while the local feature vectors $\h_j$ are obtained by applying a BN layer after $\g_j$ 
Triplet and center losses are finally applied to $\G$ and cross-entropy loss applied to $\h_j$.
The model trained with loss $L = L_{triplet} + \beta L_{center} + L_{cross}$. }
\label{fig:model}
\end{figure*}

Different loss functions have also been presented to boost the performance of the PReID models.
Two loss functions are widely used: triplet loss \cite{FaceNet} and cross-entropy loss \cite{wen2016discriminative}. 
Triplet loss is based on the feature metrics distances while cross-entropy loss is based on classification with fully connected (FC) layers.
Hermans \etal \cite{Hermans2017} and Zhang \etal \cite{zhang2019learning} modified triplet loss to increase the training performance.
Fan \etal \cite{Fan2019} presented a classification model based on an extended version of the cross-entropy loss function and a warming-up learning rate to learn a hypersphere manifold embedding. 
Recently, several models \cite{he2019mfbn, Quan2019} are trained using a combination of triplet loss and cross-entropy loss.

In this paper, we propose a multi-resolution overlapping stripes (MROS) model by combining global and local information at multiple resolutions with different loss functions.
First, based on the residual network (ResNet50) \cite{Resnet}, multiple levels are created each of which has different resolution.
Inspired by the PCB model \cite{Sun2018}, the feature map from each level is divided horizontally into multi-stripes which will be processed later in pairs with overlapping rather than individually.
The overlapping avoids lost of information
at the boundaries/edges of stripes which usually occurs when using the part-based models.
Secondly, instead of using the features from all multi-resolution levels for classification as in \cite{Wang2018}, only the features from the last two levels are considered. This is because the later levels of the model learn 
more semantic representations
compared to the early layers. 
Thirdly, local and global features are combined using different loss functions.
Experiments on the \textbf{Market-1501} \cite{market1501} dataset -- a large-scale person dataset most widely used for the PReID task -- show 
the effectiveness of the presented approach.

\section{Multi-Resolution Overlapping Stripes Model}
\label{sec:model}
Given a collection of images divided into query, gallery and training sets, PReID aims to find the 
images of each pedestrian from a query set in the gallery set.
To address this problem, we propose a multi-resolution overlapping stripes  \textbf{(MROS)} model as shown in Fig. \ref{fig:model}. 

The MROS model is constructed as follows.
Firstly, inspired by the model presented in DaReNet \cite{Wang2018}, we construct multi-level features model. 
Instead of using every feature level, we only use the last two feature levels (Section \ref{ssec:network}).
This reduces the computational complexity and increases the model performance.
Secondly, the local features are extracted by extending the PCB \cite{Sun2018} network. 
Instead of using $s$ stationary and non-overlapping stripes, an overlapping partitioning technique is employed based on pairs of stripes rather than individual ones (Section \ref{ssec:network}). 
This  technique helps our method to avoid missing features at the boundaries of the individual stripes.
Lastly, inspired by the recent successful performances achieved by local and global feature fusion \cite{Li2018, li2019pedestrian, he2019mfbn} 
and loss function fusion \cite{Luo_2019, he2019mfbn, Quan2019}, various loss functions based on local and global features are employed in this work to boost the performance of the model (Section \ref{ssec:loss}). 

%

\subsection{Network Architecture} \label{ssec:network}
As shown in Figure~\ref{fig:model}, the backbone network for our model is the ResNet-50.
It is a CNN trained on more than a million images from the ImageNet \cite{imagenet_cvpr09} database and consists of four convolutional blocks $T_{i}$, where $i=\left\{ 1,\ldots,4\right\}$.
To build a multi-resolution model, we only consider the output of the last two conv blocks, \ie tensors $T_3$ and $T_4$.
The part-based model is constructed by dividing feature tensors $T_3$ and $T_4$ to $s=6$ equal stripes, then the adjacent stripes are grouped in pairs and the global average pooling (GAP) is applied on overlapping stripes. 
For each tensor $T_j$ with $j= 3$ or $4$, the GAP operation generates new feature vectors $\g_j$ with $s-1$, $5$, stripes in each.
After that, batch normalization (BN) layers are applied on $\g_j$ to obtain $\h_j$ to overcome the overfitting and boost the performance of the system. 

For classification, FC layers are added after the local features $\h_j$. 
Note that FC layers for $\h_4$ are 2048-dimensional while FC layers for $\h_3$ are 1024-dimensional.
Feature descriptor $\G$ is defined by concatenating the feature vectors $\g_j$.
Feature vectors $\G$ and $\h_j$ are used during the training while feature vector $\G$ is used at testing.

\subsection{Loss Functions} \label{ssec:loss}
During the training, the MROS model is optimized by minimizing the fusion of three different loss functions including the triplet loss combined with center loss for metric learning and the cross-entropy loss for classification.

Firstly, instead of calculating individual losses for each stripe, a global feature $\G$ is defined by concatenating feature vectors $\g_3$ and $\g_4$.
The batch-hard triplet loss \cite{Hermans2017} is then applied on the feature vector $\G$ as follows:
\setlength{\abovedisplayskip}{1pt}
\setlength{\belowdisplayskip}{1pt}
\begin{align} \label{eq:triplet}
L_{triplet} = \sum_{i=1}^{P} \sum_{a=1}^{K} [\alpha +& \max\limits_{p=1..K} \| \G_a^i - \G_p^i \|_2	\\
- &\min_{\substack{j=1..K\\n=1..K\\i\neq j} }\| \G_a^i - \G_n^j \|_2 ]_+ \nonumber
\end{align}
where $P$ is the number of identities in a batch, $K$ is the number of images for the same identities in a batch, $\alpha$ is loss margin,
$\G_a$, $\G_p$ and $\G_n$ are features vectors from anchor, positive and negative samples. 

At this stage, the center loss \cite{wen2016discriminative} is also applied on global feature vector $\G$ to minimize the feature distribution in the feature space as following:
\begin{equation} \label{eq:center_loss}
L_{center} = \dfrac{1}{2} \sum_{i=1}^{m}
\| \G^{i} - c_{y_i} \|_2^2 
\end{equation}
where $m$ is the batch size, $c_{y_i} $ is ${y_i}$th class center vector for the features.

Secondly, the cross-entropy loss is computed for each stripe of the local feature vectors $\h_j$ as follows:
\begin{equation} \label{eq:softmax_loss}
L_{cross} = \sum_{i=1}^{m}
log \dfrac{e^{W_{y_i}^T \h^i_j + b_{y_i}}}
{\sum\limits_{c=1}^{C} e^{W_{c}^T \h^i_j + b_{c}}} 
\end{equation}
where $m$ is the batch size, $C$ is the number classes in the training set, $W$ is the weight vector for the FC layers and $b$ is the bias. Also, total cross-entropy loss is calculated as mean of all cross-entropy losses as follows:
\begin{equation} \label{eq:total_softmax_loss}
L_{cross}^{'}=\dfrac{1}{2*(s-1)} \sum\limits_{j=3}^{4}\sum\limits_{l=1}^{s-1}L_{cross}^{j,l}
\end{equation}
The label smoothing (LS) \cite{Szegedy2015RethinkingTI} technique is applied to improve the accuracy and prevent classification overfitting.


Finally, the total loss function is calculated by the weighted sum of the previous losses in Equations \ref{eq:triplet}, \ref{eq:center_loss} and \ref{eq:total_softmax_loss}, where the priority of triplet loss and cross-entropy loss are kept equal.
\begin{equation} \label{eq:loss}
\begin{split}
L = L_{triplet} + \beta L_{center} + L_{cross}^{'} 
\end{split}
\end{equation}
where $\beta$ is the weight of center loss. 

\section{Experiments} 
\label{sec:exp}
The MROS is evaluated 
using the \textbf{Market-1501} \cite{market1501}, which is a large-scale person dataset most widely used for PReID. It is collected from six different cameras with overlapping fields of view where five cameras have $1280 \times 1080$ HD resolution and one camera has $720 \times 576$ SD resolution.
The dataset has $32,668$ bounding boxes generated using a person detector for $1,501$ individuals.
Following \cite{market1501}, the dataset is split into $12,936$ images for training and $19,732$ images for testing. Single-query mode is used for searching the query images in gallery set individually.

The mean average precision (mAP) \cite{market1501}, Rank-1, Rank-5 and Rank-10 accuracies are used to evaluate the MROS performance.
The area under the Precision-Recall curve also known as average precision (AP) is calculated for each query image.
The mean value of APs over all queries is then calculated as mAP.


\subsection{Experimental Setup}
We use two Nvidia GeForce GTX $1080$ Ti GPUs with $3584$ CUDA cores and $11$ GB video memory for implementation. 
All implementations are done on Python $3.5$ with PyTorch \cite{paszke2017automatic} library.

Data augmentation is used to overcome the overfitting by artificially enlarging the training samples with class-preserving transformations.
This helps to produce more training data and reduce overfitting. 
In our experiment, different types of data augmentation are employed including zero padding with $10$ pixels, random cropping, horizontal flipping with $0.5$ probability and image normalization with the same mean and standard deviation values as ImageNet dataset \cite{imagenet_cvpr09}. Random erasing \cite{Zhong2017re} is also applied with $0.5$ probability and ImageNet pixel mean values.

We use the Adam method \cite{Kingma2014} as our optimizer using the warm-up learning rate technique \cite{Fan2019} with $0.01$ coefficient and $10$ period epochs.
The learning rate is set to  $0.001$ and is reduced using the staircase function by a factor of $0.1$ after every $30$ epochs. 
The batch size is $128$ while $P$ and $K$ in Equation~\ref{eq:triplet} are set to $32$ and $4$, respectively.
The weight of center loss $\beta$ in Equation~\ref{eq:loss} is set to $0.0005$.
The ECN \cite{ecn} is used as re-ranking method.
\subsection{Experimental Results}
\label{sec:results}

This section presents the performance evaluation of 
different settings of the  
MROS model on Market-1501 dataset. It also includes comparisons with the state-of-the-art methods.

To evaluate the effectiveness of each step of the presented model, we incrementally 
measure the accuracy as follows. 
\begin{itemize}
\item \emph{Setting \RNum{1}} presents the baseline model constructed using part-based features followed by none-overlapping stripes method with $s=6$ stripes to generate local feature vectors $\g_4$ and $\h_4$.
During this experiment, all loss functions --  triplet, center, and cross-entropy losses -- are applied on local feature vectors $\g_4$ and $\h_4$.

\item \emph{Setting \RNum{2}} is similar to \emph{Setting \RNum{1}} except that it uses overlapping stripes.


\item \emph{Setting \RNum{3}} evaluates the effectiveness of combining global and local features by generating the global feature vector $\G$ and using it with the triplet and center losses while using cross-entropy loss with $\h_j$. 

\item \emph{Setting \RNum{4}} evaluates the effectiveness of the multi-level features by considering last two level features, \ie $\g_3$ and $\g_4$. 
\end{itemize}

\begin{table}[!htbp]
\centering
\begin{tabular}{| c | c || c | c | c | c |}
\hline
\# & MROS Settings & mAP & Rank-1 \\\hline\hline
\RNum{1} & $s$ Non-Overlapping Stripes	&  81.8 	&  93.2		\\\hline 
\RNum{2} & $s-1$ Overlapping Stripes (OS)        	& 82.8 	& 93.5	\\\hline
\RNum{3} & OS with Global Features, $\G$	& 84.0	& 94.2	\\\hline
\textbf{\RNum{4}} & \textbf{Complete MROS}	& \textbf{84.2} & \textbf{94.4} \\\hline
\end{tabular}
\caption{Comparison of proposed three approaches on Market-1501
\cite{market1501} dataset.
}
\label{table:inner_result}
\end{table}

Table \ref{table:inner_result} presents these settings along with experimental results. The baseline \emph{Setting \RNum{1}} achieves promising results, however, \emph{Setting \RNum{2}} increases the performance by using overlapping stripes.
On the other hand, using global and local features boosts the performance in \emph{Setting \RNum{3}}.
Finally, the best results are obtained by \emph{Setting \RNum{4}} which combines all previous settings with multi-resolution features.

\begin{table}[!htbp]
\centering
\small
\scalebox{0.96}{
\begin{tabular}{| c || c | c | c | c |}
\hline
Model 						& mAP 	& Rank-1 & Rank-5 & Rank-10 \\\hline\hline
DaRe \cite{Wang2018}		& 76.0	& 89.0	& 	- 	&	-	\\
HA-CNN \cite{Li2018}		& 75.7	& 91.2	&	- 	&	-	\\
PCB  \cite{Sun2018}			& 77.4 	& 92.3	& 97.2 	& 98.2	\\
TBN+ \cite{li2019pedestrian}& 83.0	& 93.2  &	-	&	-	\\
PCB+RPP \cite{Sun2018}		& 81.6 	& 93.8	& 97.5 	& 98.5	\\
MFBN \cite{he2019mfbn}		& 84.9	& 93.9	&	-	&	-	\\
SphereReID  \cite{Fan2019}	& 83.6	& 94.4 	& 98.0 	& 98.7	\\
Auto-ReID  \cite{Quan2019}	& 85.1	& 94.5 	& 98.5 	& 99.0	\\
Strong ReID \cite{Luo_2019} & 85.9	& 94.5 	& 	- 	& 	-	\\
\hline
\textbf{Proposed MROS}				& \textbf{84.2 }	& \textbf{94.4}	& \textbf{97.8}	& \textbf{98.7} \\\hline\hline

\end{tabular}
}
\caption{Comparison with the state-of-the-art results on the Market-1501 \cite{market1501} dataset without re-ranking.} 
\label{table:sota_result}
\end{table}

A comparison of the experimental results between MROS 
using single-query mode and the related methods are presented in Table~\ref{table:sota_result} and Table~\ref{table:sota_result_rr} without and with re-ranking, respectively.
The MROS model achieved \textbf{mAP = 84.2\%  and Rank-1 = 94.4\%} without re-ranking and \textbf{mAP = 93.5\%  and Rank-1 = 95.5\%}
with re-ranking \cite{ecn}. 
The results in Table~\ref{table:sota_result} show that the proposed MROS model without re-ranking achieves competitive performances. 
On the other hand, most of the re-ranked PReID models in Table~\ref{table:sota_result_rr} reported rank-1 results in a small margin $[95.1$-$95.4]$. As can be observed from the table, our MROS model outperforms the state-of-the-art models.
This is because MROS is more able to learn body parts and the spatial correlation between them by employing the overlapping stripes and learn discriminative features by employing multi-resolution and different loss functions.

\begin{table}[!htbp]
\centering
\small
\scalebox{0.96}{
\begin{tabular}{| c || c | c | c | c |}
\hline
Model 						& mAP 	& Rank-1 & Rank-5 & Rank-10 \\\hline\hline
%
DaRe \cite{Wang2018}		& 86.7	& 89.0	& 	- 	&	-	\\
GSRW  \cite{GSRW} 			& 82.5 	& 92.7	& 96.9 	& 98.1	\\
PCB+RPP \cite{Sun2018} 		& 81.9 	& 95.1	& 	- 	& 	-	\\
MFBN \cite{he2019mfbn}		& 93.2	& 95.2	& 	-	& 	- 	\\
TBN+ \cite{li2019pedestrian}& 91.3	& 95.4  &	-	&	-	\\
Auto-ReID  \cite{Quan2019}	& 94.2	& 95.4	& 97.9 	& 98.5	\\
Strong ReID  \cite{Luo_2019}& 94.2	& 95.4 	& - 	& 	-	\\ \hline
\textbf{Proposed MROS }		& \textbf{93.5}	& {\textbf{95.5}}	& \textbf{97.2}	&\textbf{ 97.8}	\\\hline\hline

\end{tabular}
}
\caption{Comparison with the state-of-the-art results on the Market-1501 \cite{market1501} dataset with re-ranking.} 
\label{table:sota_result_rr}
\end{table}

\section{Conclusions} 
\label{sec:conclusion}
This paper extended the part-based convolutional baseline (PCB) and the multi-resolution model to solve the problem of pedestrian retrieval. 
Using the residual network (ResNet50) as backbone network, multi-levels with different resolutions are created to generate feature maps.
After that a simple uniform partition technique is applied on the last two 
conv blocks and the generated features are combined with overlapping. 
Using different types of loss functions, both global and local representations were considered for classification. 
Experimental results show that our approach outperforms the state-of-the-art methods.
\bibliographystyle{IEEEbib}
\bibliography{refs}

\end{document}